\documentclass[10pt,twocolumn,letterpaper]{article}

\usepackage[pagenumbers]{iccv} % To force page numbers, e.g. for an arXiv version

\usepackage{algorithm}
\usepackage{algpseudocode} 
\usepackage{amsmath}
\usepackage{amssymb}
\usepackage{amsthm}
\usepackage{array}
\usepackage{bbding}
\usepackage{booktabs}
\usepackage{CJKutf8} 
\usepackage{colortbl}  % 颜色表格支持
\usepackage{graphicx}
\usepackage{grffile}

\usepackage{makecell}
\usepackage{multirow}
\usepackage{newfloat}
\usepackage[numbers,sort&compress]{natbib}
\usepackage{pifont}
\usepackage{romannum}
\usepackage{soul}
\usepackage[switch]{lineno}
\usepackage{tabularx}
\usepackage{times}
\usepackage{todonotes}

\usepackage{url}
\usepackage[utf8]{inputenc}

\usepackage{xargs} 
\usepackage{xspace}

\definecolor{lightred}{RGB}{255, 200, 200}  % 浅红色
\definecolor{lightblue}{RGB}{200, 220, 255} % 浅蓝色
\definecolor{mycolor}{rgb}{0.902,0.902,0.980}
\definecolor{darkgreen}{rgb}{0.2,0.8,0.25}

\definecolor{iccvblue}{rgb}{0.21,0.49,0.74}
\usepackage[pagebackref,breaklinks,colorlinks,allcolors=iccvblue]{hyperref}

\def\paperID{2284} % *** Enter the Paper ID here
\def\confName{ICCV}
\def\confYear{2025}

\title{MultiTaskVIF: Segmentation-oriented visible and infrared image fusion via multi-task learning}

\author{Zixian Zhao$^1$, Andrew Howes$^2$, Xingchen Zhang$^{1*}$\\
$^1$The Fusion Intelligence Laboratory, Department of Computer Science, \\ University of Exeter, EX4 4QF, Exeter, UK\\
$^2$Department of Computer Science, University of Exeter, EX4 4QF, Exeter, UK\\
{\tt\small \{zz541, andrew.howes, X.Zhang12\}@exeter.ac.uk}
}

\begin{document}
\maketitle

\begin{abstract}
Visible and infrared image fusion (VIF) has attracted significant attention in recent years.~Traditional VIF methods primarily focus on generating fused images with high visual quality, while recent advancements increasingly emphasize incorporating semantic information into the fusion model during training.~However, most existing segmentation-oriented VIF methods adopt a cascade structure comprising separate fusion and segmentation models, leading to increased network complexity and redundancy. This raises a critical question: can we design a more concise and efficient structure to integrate semantic information directly into the fusion model during training?~Inspired by multi-task learning, we propose a concise and universal training framework, \textbf{MultiTaskVIF}, for segmentation-oriented VIF models.~In this framework, we introduce a multi-task head decoder (MTH) to simultaneously output both the fused image and the segmentation result during training. Unlike previous cascade training frameworks that necessitate joint training with a complete segmentation model, MultiTaskVIF enables the fusion model to learn semantic features by simply replacing its decoder with MTH.~Extensive experimental evaluations validate the effectiveness of the proposed method. Our code will be released upon acceptance.
\end{abstract}
\section{Introduction}
\begin{figure}[ht]
	\centering
	\includegraphics[width=0.5\textwidth]{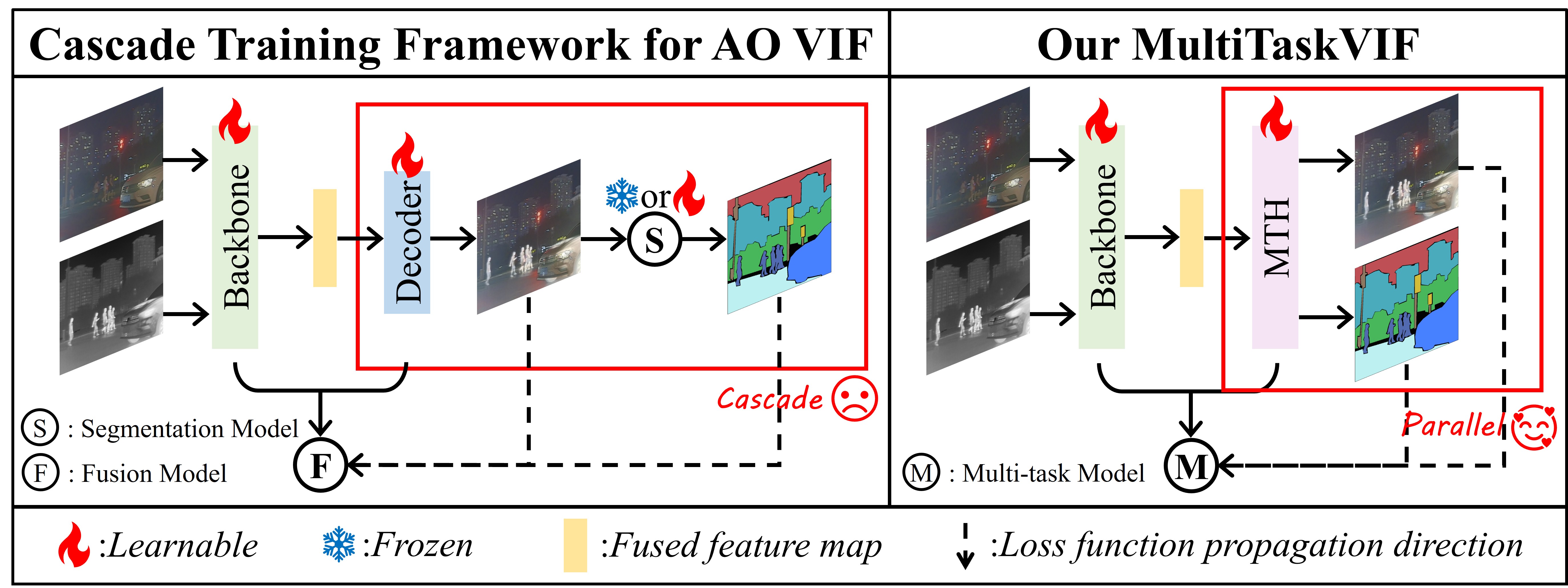}
	\caption{Existing cascade training framework vs. MultiTaskVIF. Both frameworks can be used to train segmentation-oriented VIF models. However, our MultiTaskVIF, which incorporates a multi-task head (MTH), is more concise as it eliminates the need for an additional full segmentation model.}
	\label{fig:1}
\end{figure}

Visible and infrared image fusion (VIF) \cite{ma2019infrared,zhang2023visible,zhang2024mrfs} combines complementary information from visible and infrared images to generate a fused image, a field that has attracted intensive attention recently.~Generally, VIF has two primary objectives.~The first is to generate visually pleasing fused images for human observation.~The second is to enhance the performance of downstream applications, such as object tracking \cite{zhang2019object}, object detection \cite{liu2022target}, and semantic segmentation \cite{tang2022image}, by providing high-quality fused images as input.

Many VIF methods \cite{li2019densefuse, tang2022superfusion, tang2022ydtr, rao2023gan, yue2023dif, zhang2023transformer, zhao2023cddfuse} have been proposed, which can be generally categorized into conventional and deep learning-based approaches. Among deep learning-based methods, the main types include CNN-based, GAN-based, autoencoder-based, transformer-based, and diffusion model-based methods.~However, most previous studies primarily focus on the first objective of VIF, i.e., generate visually pleasing fused images.~In other words, downstream applications are not considered in those VIF methods, resulting in a gap between the VIF method and its practical use.~Consequently, their effectiveness in specific downstream application tasks remains uncertain.

In recent years, some researchers \cite{zhang2023visible, tang2022image, liu2022target, liu2023segmif} have recognized this gap and  have sought to achieve the two objectives of VIF concurrently.~For example, Shopovska et al.~\cite{shopovska2019deep} proposed to use an auxiliary detection network following the image fusion network during training, aiming to enhance pedestrian visibility in fused images.~Tang et al.~\cite{tang2022image} incorporated semantic segmentation into the training process of image fusion model.~Liu et al.~\cite{liu2022target} trained the image fusion and object detection model simultaneously by a proposed joint training approach.~However, as shown in Fig.~\ref{fig:1}, most existing application-oriented (AO) VIF methods employ a cascade structure for the training process, where an image fusion model is followed by a %cascaded 
downstream application model. In this cascade training framework, the fusion model's output will be fed into the application model, which in turn provides feedback to refine the training of the fusion model. By using this framework, semantic information can be incorporated into the fusion model and thus improving its downstream application performance.

Although current AO VIF methods have demonstrated outstanding performance in various downstream tasks \cite{liu2023segmif, sun2022detfusion}, their cascade training framework has three main shortcomings. First, the cascade training framework will introduce potential redundancy of models. For example, image feature extraction is performed separately in both the image fusion and downstream application models. Second, the cascade frameworks often adopt multi-stage training strategies \cite{tang2022image, liu2023segmif, liu2023paif}. For example, in Tang et al. \cite{tang2022image}, the fusion model is trained for $p$ steps, and then the fused dataset is used to train the cascaded segmentation model for next $q$ steps. Compared to single-stage training, multi-stage training requires more careful hyperparameter tuning for each stage, increasing the overall training efforts. It may also introduce potential cumulative errors, as early-stage errors may propagate to later stages \cite{xu2022advanced}. Third, the cascade frameworks will increase memory usage since multiple models require more parameters and memory than a single model. Even during inference, some AO VIF methods must run both the fusion and joint trained application models simultaneously \cite{liu2023segmif}, leading to inefficient processing and further intensifying memory pressure.

In this paper, to address the above-mentioned issues, we propose \textbf{MultiTaskVIF}, a universal segmentation-oriented VIF training framework that is more concise and efficient than previous cascade frameworks, as shown in Fig.~\ref{fig:1}.~Inspired by the characteristic of multi-task learning methods \cite{kendall2018multi}, where multiple task heads can share the same backbone for feature extraction, we designed \textbf{a multi-task head (MTH)} that takes fused feature as input and outputs both the reconstructed fused image and scene segmentation results.~In our MultiTaskVIF training framework, we replace the decoder in previous image fusion models with designed MTH and leverage MTH's segmentation output to assist the overall fusion model training. Consequently, in our parallel training framework, there is no need to introduce an additional segmentation model to provide semantic information for the fusion model, thereby reducing the memory pressure during training. Moreover, by allowing the segmentation head and fusion reconstruction head within MTH to share the same fusion backbone, we eliminate the model redundancy issues inherent in previous cascade training frameworks. Training a single model is also significantly easier and more straightforward compared to multi-stage training involving multiple models.

In summary, the main contributions of this paper include:
\begin{itemize}

    \item We propose a universal parallel training framework for AO VIF methods: \textbf{MultiTaskVIF}. Unlike previous redundant and complex cascade frameworks, this framework enables VIF models to integrate more semantic segmentation information during training in a more concise and efficient manner, offering a novel alternative for training segmentation-oriented VIF model.

    \item We propose a \textbf{M}ulti-\textbf{T}ask \textbf{H}ead (\textbf{MTH}) designed based on the concept of multiple task branches sharing the same backbone in multi-task learning models. This module can flexibly replace the decoder in previous VIF models and effectively enhance the fusion backbone's ability to learn semantic information.

    \item Extensive experiments demonstrate the effectiveness of our MultiTaskVIF. Compared to cascade training frameworks, MultiTaskVIF achieves a more concise and efficient improvement in both image fusion and segmentation performance of fused images, using only a single model and training stage. Moreover, it is compatible with %most 
    advanced VIF methods, demonstrating its generalizability.
    
\end{itemize}

\section{Related work}
\subsection{Visible and infrared Image fusion (VIF)}
With the development of deep learning (DL), DL-based image fusion and VIF \cite{zhang2023visible, ma2019fusiongan, xu2020u2fusion} methods have attracted increasing attention from researchers in recent years. To the best of our knowledge, Liu et al. \cite{liu2018infrared} conducted one of the earliest studies on applying deep learning to the VIF task \cite{zhang2023visible}. They proposed using a Siamese CNN to generate a weight map from source images, fusing visible and infrared images in a multiscale manner.~%DenseFuse \cite{li2019densefuse}, a well-known autoencoder-based VIF method, uses the MS-COCO dataset \cite{lin2014microsoft} to pretrain the autoencoder, with fusion strategies (addition and $l_1$-norm) used to enhance VIF performance.
In recent years, models like GANs, transformers, and diffusion models have become common tools for the VIF task. For example, Yue et al. \cite{yue2023dif} introduced a diffusion model for multi-channel feature extraction to preserve color features during VIF, resulting in better fused images with improved fusion performance.~However, most of these VIF methods do not consider downstream applications in the image fusion process.

\begin{figure*}
	\centering
	\includegraphics[width=0.83\textwidth]{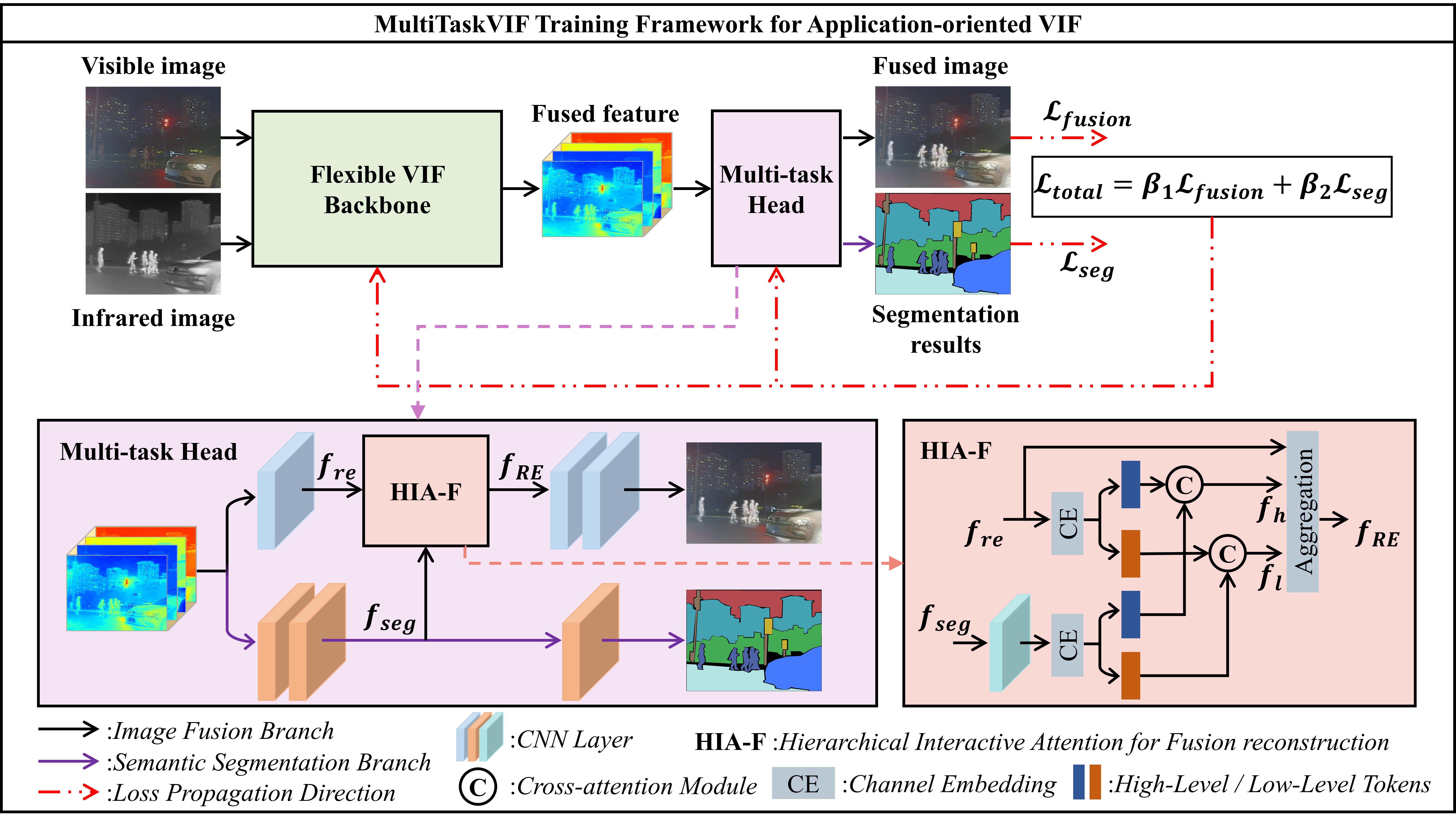}
	\caption{The workflow of proposed MultiTaskVIF training framework for AO VIF. Notably, MultiTaskVIF exhibits strong generalizability and is compatible with most existing VIF methods \cite{ma2022swinfusion, zhao2024equivariant, liu2023segmif, tang2022image}. Moreover, through this network architecture and training strategy, it effectively enhances the performance of VIF models in both image fusion and semantic segmentation tasks.}
	\label{fig:idea}
\end{figure*}

\subsection{Application-oriented (AO) VIF methods}
Since 2019, several AO VIF methods have been proposed.~For instance, SeAFusion \cite{tang2022image} is the first VIF method to consider segmentation performance of fused images during VIF training.~It cascaded a learnable segmentation model after the fusion model, and jointly trained these two model.~SegMiF \cite{liu2023segmif} is another segmentation-oriented VIF method which introduces a hierarchical interactive attention module to integrate additional semantic information from cascaded segmentation model to the modality features. Its fusion and segmentation models are connected sequentially, resulting in two separate encoding-decoding processes with feature interaction.~Similarly, IRFS \cite{wang2023interactively} is a cascade framework with an image fusion model and a salient object detection (SOD) model, enhancing SOD performance by incorporating SOD loss during training.~According to our literature review, most existing AO VIF methods adopt the cascade training framework, as illustrated in Fig. \ref{fig:1}.

However, in recent years, there has been growing exploration of new training frameworks for AO VIF methods. For example, MRFS \cite{zhang2024mrfs}, published in 2024, adopts a multi-task learning framework to jointly train the fusion model and the segmentation model. However, their approach focuses on separately improving fusion and segmentation performance of fusion and segmentation branch, which contradicts the initial objective of AO VIF methods that aim to improve the downstream application performance of the fused image. Nevertheless, MRFS remains an inspiring and valuable work, as it demonstrates the feasibility of multi-task learning frameworks in the VIF domain.

\subsection{Multi-task learning (MTL)}
Multi-task learning \cite{ruder2017overview} is an important area in machine learning, widely applied to tasks such as computer vision \cite{girshick2015fast}, natural language processing \cite{collobert2008unified} and speech recognition \cite{deng2013new}.%, and drug discovery \cite{ramsundar2015massively}.~of 
~In computer vision, multi-task learning is often used to learn strongly correlated tasks, such as cross-domain image classification \cite{rebuffi2017learning}, pose estimation and action recognition \cite{gkioxari2014r}, object recognition and object detection \cite{sermanet2013overfeat}, as well as depth estimation and semantic segmentation \cite{xu2023demt, kendall2018multi}.~Very few multi-task learning studies have attempted to integrate VIF and semantic segmentation into a single multi-task model.~MRFS \cite{zhang2024mrfs} is an inspiring exploration, which explicitly proposes the use of an MTL model to jointly learn VIF and semantic segmentation tasks, demonstrating the strong correlation between VIF task and semantic segmentation task. However, it is worth noting that our work differs significantly from traditional MTL. While MTL emphasizes mutual enhancement between tasks and the generalizability of a multi-task model across multiple tasks \cite{kendall2018multi}, our work focuses solely on the model's performance in the VIF task, without considering utilizing our model for segmentation task. The segmentation output of our multi-task head is only used to assist the training of the fusion model.

\section{Method}

\subsection{Problem formulation}
In this study, as shown in Figs. \ref{fig:1} and \ref{fig:idea}, we propose incorporating additional semantic supervision into VIF models through the structure of a multi-task model.

We assume the visible, infrared, and fused images are all 3-channel RGB images of size $H \times W$, represented as $v$, $i$, and $f\in \mathbb{R}^{3 \times H \times W}$. We propose that the segmentation task can share the backbone of image fusion task, thereby reducing model redundancy and training complexity found in previous cascade training framework \cite{liu2023segmif, tang2022image}. Based on this motivation, our AO VIF model's optimization problem can be formulated as:

\begin{equation}
\min_{\omega_b, \omega_{f}, \omega_{s}} \mathcal{F}_f\left(f, \mathcal{N}_m(v, i; \omega_b, \omega_{f})\right) + \mathcal{F}_s\left(s, \mathcal{N}_m(v, i; \omega_b, \omega_{s})\right),
\end{equation}
where $\mathcal{F}(\cdot)$ denotes a fidelity term, $\mathcal{N}_m$ represents the multi-task model, and $s$ denotes the segmentation result. $\omega_b$ refers to the shared learnable parameters of the backbone in $\mathcal{N}_m$, while $\omega_{f}$ and $\omega_{s}$ correspond to the learnable parameters of the fusion reconstruction and semantic segmentation branches in MTH, which will introduced in Section \ref{Sec.MTH}.~Therefore, the optimization of our AO VIF model reduces to optimizing a single multi-task model, where the total number of parameters is given by $\omega_b+\omega_f+\omega_s$.

\subsection{MultiTaskVIF training framework}
\label{Sec.MultiTaskVIF}
\noindent\textbf{Workflow.}~As shown in Fig. \ref{fig:idea}, during our proposed training process, visible and infrared images are simultaneously fed into a flexible VIF backbone for feature extraction and fusion.~The deep fused features generated by the backbone are then processed by MTH, which produces both the fused image and the segmentation result. These outputs provide their respective task-specific loss values, $\mathcal{L}_{fusion}$ and $\mathcal{L}_{seg}$, which together form the overall loss, $\mathcal{L}_{total}$. Finally, $\mathcal{L}_{total}$ is used to supervise the parameter updates of the entire multi-task model.

\noindent\textbf{Flexible backbone.}~Since most DL-based VIF methods have backbones that involve both feature extraction and feature fusion, our MultiTaskVIF is not limited to a specific VIF method but rather serves as a general training framework for AO VIF models. In the subsequent experiments in Section \ref{sec:exp}, we trained MultiTaskVIF using the backbones of two pure VIF methods \cite{ma2022swinfusion, zhao2024equivariant} and two AO VIF methods \cite{tang2022image, liu2023segmif}. All four models demonstrated superior fusion and segmentation performance compared to their original methods, further validating the generalizability of our proposed MultiTaskVIF training framework.

\noindent\textbf{Decoder.}~One of the core innovations of MultiTaskVIF lies in the modification of the decoder in previous VIF methods. Unlike previous methods, where the decoder produces a single output, we replace the original decoder with our designed MTH, enabling the simultaneous output of both the fused image and segmentation results. The segmentation output provides a direct supervision signal for semantic information (e.g., segmentation labels), allowing the integration of semantic information into VIF model through properly designed loss functions.

It is worth noting that we describe MultiTaskVIF as providing direct semantic supervision for VIF training because the segmentation branch is updated as an integral part of the model and directly influences both the fusion reconstruction branch and the backbone training. In contrast, previous cascade training frameworks keep the segmentation model independent from the fusion model, making their semantic supervision indirect. Subsequent experiments also demonstrate that our direct semantic supervision in MultiTaskVIF is more effective in guiding the training of AO VIF models than previous cascade training frameworks.

\subsection{Multi-task Head (MTH)}
\label{Sec.MTH}
\noindent\textbf{Dual-branch structure.}~Inspired by multi-task learning studies \cite{kendall2018multi, sermanet2013overfeat}, we design a dual-branch decoder, the \textbf{M}ulti-\textbf{T}ask \textbf{H}ead (\textbf{MTH}). As shown in Fig. \ref{fig:idea}, MTH consists of two branches: fusion reconstruction branch and semantic segmentation branch. Both branches take the same deep fused features as input and, through a series of convolutional neural network (CNN) layers, separately accomplish the tasks of fused image reconstruction and segmentation.

\noindent\textbf{HIA-F module.}~To facilitate the integration of semantic information into the fusion model, we design a \textbf{H}ierarchical \textbf{I}nteractive \textbf{A}ttention module for \textbf{F}usion reconstruction (HIA-F), building on previous work of HIA \cite{liu2023segmif}. Unlike HIA, which integrates individual modality features and semantic features during the feature extraction stage, our proposed HIA-F enables interaction between semantic features and fusion features during the fusion reconstruction stage.

In detail, we first use a CNN layer~%to further extract the semantic segmentation feature $f_{seg}$ and 
to align the semantic segmentation feature $f_{seg}$ with the fusion reconstruction feature $f_{re}$. Then, channel embedding is applied to decompose $f_{seg}$ into tokens ${t_{s}^{h},t_{s}^{l}}$, where $t_{s}^{h}$ and $t_{s}^{l}$ represent the high-level and low-level features of $f_{seg}$. For $f_{re}$, it is also decomposed into ${t_{r}^{h},t_{r}^{l}}$ through channel embedding, with each token capturing different level characteristics. Then, we input ${t_{s}^{h},t_{r}^{h}}$ and ${t_{s}^{l},t_{r}^{l}}$ into the cross-attention modules $C_h$ and $C_l$, respectively, to facilitate feature interaction. In $C_h$, $t_{s}^{h}$ serves as the query, while $t_{s}^{l}$ serves as the key and value, filtering out the features in $f_{re}$ that are more useful for high-level tasks, resulting in $f_h$. In contrast, in $C_l$, $t_{r}^{l}$ serves as the query, while $t_{s}^{l}$ serves as the key and value, extracting features from $f_{seg}$ that are more beneficial for low-level tasks, yielding $f_l$. Finally, in the aggregation layer, we concatenate $f_h$ and $f_l$ and pass them through an MLP layer with a residual connection to $f_{re}$, generating $f_{RE}$. The resulting $f_{RE}$ not only retains the original features of $f_{re}$ but also incorporates low-level features from $f_{seg}$ beneficial for fusion reconstruction while enhancing the high-level features in $f_{re}$ that support semantic segmentation.~Thus, through the above method, \textit{our proposed HIA-F enables comprehensive information interaction between semantic segmentation features and fusion reconstruction features}.~This further enhances the fusion model trained with MultiTaskVIF, allowing it to generate fused images with higher fusion quality and greater utility for segmentation tasks.

\subsection{Loss functions}
Our loss function consists of two parts, i.e., image fusion loss and segmentation loss:
\begin{equation}
\label{eq.total}
\mathcal{L}_{total} = \beta_1 \mathcal{L}_{fusion} + \beta_2 \mathcal{L}_{seg},
\end{equation}
where $\mathcal{L}_{fusion}$ is the image fusion loss and $\mathcal{L}_{seg}$ is the application loss, $\beta_1$ and $\beta_2$ %are parameters to 
control the ratio of two loss terms.

\noindent\textbf{Image fusion loss.}%\label{subsubsec:vif-loss}.
~In order to retain sufficient pixel intensity, texture, and structural information in the fused image, we designed intensity loss $\mathcal{L}_{int}= \frac{1}{HW} \mathbb{E}_i \| I_f^i - \max(I_{ir}^i, I_{vis}^i) \|_1$, gradient loss $\mathcal{L}_{grad} = \frac{1}{HW} \mathbb{E}_i \| |\nabla I_f^i| - \max( |\nabla I_{ir}^i|, |\nabla I_{vis}^i| ) \|_1$, and structural similarity loss $\mathcal{L}_{ssim}= \sum_{j \in \{ir, vis\}} w_j \cdot \left( 1 - \mathbb{E}_{i} \big[ ssim(I_f^i, I_j^i) \big] \right)$.~$I_{f}^i (i\in{1,2,3})$ represents the three channels of the fused image $I_{f}$; the same applies to the infrared image $I_{ir}$ and the visible image $I_{vis}$. $(H, W)$ denote the height and width of the input images, and $\mathbb{E}_{i}$ represents the expectation (mean) over index $i$. $\nabla$ represents the Sobel gradient operator. $\left|\cdot\right|$ and $\left\|\cdot\right\|_1$ indicate the absolute value and $l_1$-norm operations, respectively. $ssim(\cdot)$ \cite{wang2004image} represents the structural similarity measure between two images, and inspired by \cite{liu2023segmif, ma2022swinfusion}, we set the weight of each term in $\mathcal{L}_{ssim}$ to $0.5$.

Additionally, to ensure color fidelity in the fused image, we further introduce a color-preserving loss, $\mathcal{L}_{color}$, into the fusion loss.~Since the $Cb$ and $Cr$ channels in $YCbCr$ space represent color information, we transform the RGB images into $YCbCr$ space and calculate $\mathcal{L}_{Color} = \frac{1}{HW} \mathbb{E}_{c \in \{Cb, Cr\}} \| I_f^c - I_{vis}^c \|_1$, where $I_f^c$ and $I^c_{vis}$ ($c\in{Cb,Cr}$) represent the fused and visible images' $CbCr$ channels in $YCbCr$ space.

In summery, our fusion loss $\mathcal{L}_{fusion}$ contains four loss terms and is calculated as:
\begin{equation}
\label{eq.fusion}
\mathcal{L}_{fusion} = \lambda_1 \mathcal{L}_{int} + \lambda_2 \mathcal{L}_{grad} + \lambda_3 \mathcal{L}_{ssim} + \lambda_4 \mathcal{L}_{color},
\end{equation}
where $\lambda_1,\lambda_2,\lambda_3,\lambda_4$ are hyper-parameters controlling the trade-off of each sub-loss term.

\noindent\textbf{Semantic segmentation loss.}
\label{subsubsec:application-loss}
To optimize region overlap while maintaining pixel-level classification accuracy \cite{azad2023loss}, we adopt a combination of cross-entropy loss $\mathcal{L}_{ce}$ and dice loss $\mathcal{L}_{dice}$ \cite{sudre2017generalised} for the semantic segmentation loss. Specifically, let $C$ denote the number of segmentation classes, $y_{i,c}$ be a binary label ($0$ or $1$) indicating whether pixel $i$ belongs to class $c$, and $p_{i,c}$ be the predicted probability that pixel $i$ belongs to class $c$.~Then, $\mathcal{L}_{ce}$ and $\mathcal{L}_{dice}$ can be calculated as $\mathcal{L}_{ce}= -\mathbb{E}_{i, c} \, y_{i,c} \log(p_{i,c})$ and $\mathcal{L}_{dice}= 1 - \mathbb{E}_{c} \frac{2 \sum_{i} p_{i,c} \cdot y_{i,c}}{\sum_{i} p_{i,c} + \sum_{i} y_{i,c}}$. Consequently, the final semantic segmentation loss is given by $\mathcal{L}_{seg}=\mathcal{L}_{ce} + \mathcal{L}_{dice}$.

\noindent\textbf{Hyperparameter settings.}~As mentioned in Section \ref{Sec.MultiTaskVIF}, MultiTaskVIF is a general AO VIF training framework that is compatible with various VIF backbones.~Therefore, the adjustment of hyperparameters in Eqs.~\ref{eq.total} and \ref{eq.fusion} depends on the complexity of the utilized backbone and the overall model size after integrating MTH.~Inspired by existing works \cite{ma2022swinfusion, liu2023segmif} and manual tuning, the hyperparameters used in our subsequent experiments are set as $(\beta_1,\beta_2,\lambda_1,\lambda_2,\lambda_3,\lambda_4)=(1,1,20,20,10,20)$.~MultiTaskVIF and MTH provide a solid platform for further research in AO VIF, and future studies based on MultiTaskVIF could explore more effective designs for loss functions and hyperparameter tuning to further improve the performance of AO VIF models.

\section{Experiments and results}
\label{sec:exp}

\subsection{Datasets and implementation details}

\noindent\textbf{Datasets.}~We evaluate the performance of our proposed framework MultiTaskVIF on semantic segmentation and image fusion tasks using two representative multi-modality segmentation datasets: FMB \cite{liu2023segmif} and MSRS \cite{tang2022piafusion}.~FMB contains 1500 infrared and visible image pairs with 15 annotated pixel-level classes for semantic segmentation, and MSRS contains 1444 image pairs with 9 classes.~The image size of FMB and MSRS is $600\times800$ and  $480\times640$.~The number of training set in FMB and MSRS is 1220 and 1083, respectively, while the other image pairs in FMB (280) and MSRS (361) are used for evaluation.~Their segmentation labels are also used in training and evaluation.
\begin{table}[ht]
\centering
\renewcommand\arraystretch{1}
\resizebox{0.45\textwidth}{!}{
\begin{tabular}{c|cccccc}
\hline 
%\midrule[1.2pt] 
Model &Types &Backbone &Decoder&Model count&Train. stages\\
\midrule 
SwinF. & VIF & SwinF. & CNN & 1&1\\ 
EMMA & VIF & Ufuser & CNN & 3&2\\ 
 SeAF.& AO VIF & SeAF. & CNN & 2&2\\
 SegMiF& AO VIF & SegMiF & CNN & 2&2\\ \midrule 
\Romannum{1}& AO VIF & SwinF. & MTH & 1&1\\
\Romannum{2}& AO VIF & Ufuser & MTH & 1&1\\
\Romannum{3}& AO VIF & SeAF. & MTH & 1&1\\
\Romannum{4}& AO VIF & SegMiF & MTH & 1&1\\
%\midrule[1.2] 
\hline 
\end{tabular}
}
\caption{Four variants of MultiTaskVIF with different backbones.}
\label{table.Varients}
\end{table}

\noindent\textbf{Model setup.}~To demonstrate the generalizability and flexibility of our MultitaskVIF, we constructed 4 models (MultiTaskVIF \Romannum{1} to \Romannum{4}) shown in Table \ref{table.Varients} based on the backbones of 4 advanced VIF methods, each incorporating our designed MTH within a multi-task learning architecture. These models were trained from scratch on the training sets of FMB and MSRS, respectively.~It is worth noting that although we selected four VIF methods (SwinFusion \cite{ma2022swinfusion}, EMMA \cite{zhao2024equivariant}, SeAFusion \cite{liu2022target} and SegMiF \cite{liu2023segmif}), we only utilized their backbones within our framework. Our subsequent training strategies were entirely independent of these methods.~For instance, EMMA employs two additional U-Net models for auxiliary training, while SeAFusion and SegMiF incorporate segmentation models for training application-oriented fusion models, but such strategies were not adopted in our framework.

\noindent\textbf{Implementation details}.~We only have one training stage and learnable model during training.~Before training, we partitioned the provided training set of FMB and MSRS into training part and validation part in a 9:1 ratio.~All our models were trained for 100 epochs within the early stopping technique (patience was set to be 10 epochs) to avoid over-fitting.~For MultiTaskVIF \Romannum{2} to \Romannum{4} shown in Table \ref{table.Varients}, the training samples were randomly cropped into $256\times256$ patches with a batch size of 16 before being fed into the network, and the initial learning rate was set to $0.5e^{-4}$ with the Adam optimizer. MultiTaskVIF \Romannum{1} also utilized the Adam optimizer, but with different initial learning rate ($1e^{-4}$), crop size ($160\times160$) and batch size (10), due to the different complexities of backbones.~All ours models are implemented in PyTorch, with training on an NVIDIA A100 GPU and testing on a Tesla T4 GPU.

\subsection{Compared methods and metrics}
\noindent\textbf{Compared methods.}~We compared our MultiTaskVIF with SOTA methods of both VIF and application-oriented (AO) VIF methods, including CDDFuse \cite{zhao2023cddfuse} (CVPR '23), TIMFusion \cite{liu2024task} (TPAMI '24), SwinFusion \cite{ma2022swinfusion} (JAS '22), EMMA \cite{zhao2024equivariant} (CVPR '24), PAIF \cite{liu2023paif} (ACMMM '23), MRFS \cite{zhang2024mrfs} (CVPR '24), SeAFusion \cite{tang2022image} (InfFus '22), and SegMiF \cite{liu2023segmif} (ICCV '23).~All models of these methods used in our evaluation experiments were the default models provided in their publicly available code repositories.

\noindent\textbf{Semantic segmentation metrics.}~To evaluate segmentation performance of fused images, we used the Intersection-over-Union (IoU) of each classes and mean IoU (mIoU).~Since there is no bike class in FMB's test set, we ignored this class during evaluation.~It is worth noting that our work focuses on enhancing the semantic segmentation performance of fused images. Although our MultiTaskVIF framework can simultaneously generate both fused images and segmentation results, we evaluate segmentation performance based solely on the fused images. Specifically, the IoU is calculated using segmentation results obtained from a unified segmentation model applied to the fused images. This evaluation strategy is consistently applied to all compared fusion models.

\begin{table}[t]
\centering
\resizebox{0.45\textwidth}{!}{
\begin{tabular}{c|cccccc|c}
\midrule[1.2pt] 
FMB &Back.&Per. &Road &Car &Bus&Lamp&mIoU\\
\midrule 
% \multicolumn{8}{c}{Pure VIF}\\% &  &  &  &  &  &  &   \\ 
CDDF. &36.2 &68.5 & 90.2 & 83.5&74.7 &43.2  &65.1  \\  
TIMF. &31.5 &63.9 &88.2  &82.8 &59.5 &43.9  &62.3  \\
SwinF. &33.5 &69.8 &90.3  &83.2 &72.3 &42.3  &64.6  \\
EMMA &34.8 &67.9 &90.2  &83.5 &73.1 &42.8  &64.7  \\ %\midrule 
 PAIF &37.6 &66.5 &89.9  &82.6 &73.0 &42.2  &65.2  \\
MRFS &34.4 &64.8 &89.6  &82.6 &72.8 &42.6  &64.3  \\
 SeAF. &36.6 &70.2 &90.3  &83.2 &72.6 &45.5  &65.2  \\
 SegMiF &36.6 &63.6 &90.3  &83.2 &72.7 &43.4  &65.4  \\ \midrule 
 \rowcolor{lightblue}\Romannum{1} &35.8 &70.3 &90.4  &83.9 &73.0 &45.0  &\underline{65.7}  \\
\Romannum{2} &37.6 &68.3 &89.9  &83.4 &72.0 &42.5  &65.1  \\
\Romannum{3} &36.8 &70.2 &90.6  &83.9 &72.2 &42.3  &65.6  \\
\rowcolor{lightred}\Romannum{4} &36.8 &69.4 &90.3  &83.8 &73.6 &45.4  &\textbf{65.8}  \\
\midrule[1.2pt] 
\end{tabular}
}
\caption{Quantitative segmentation (IoU\%↑) on the FMB dataset. Best and second-best values are \textbf{highlighted} and \underline{underlined}.}
\label{table.FMB_segmentation}
\end{table}

\begin{table}[t]
\centering
\resizebox{0.45\textwidth}{!}{
\begin{tabular}{c|cccccc|c}
\midrule[1.2pt] 
MSRS &Back.&Car &Per. &Bike &Curve&Bump&mIoU\\
\midrule 
% \multicolumn{8}{c}{Pure VIF}\\% &  &  &  &  &  &  &   \\ 
CDDF. &98.5  &91.1  &74.5  &69.5  &59.8  &80.1  & 76.2  \\  
TIMF. &98.2  &89.1  &67.9  &65.7  &47.8  &67.8  & 71.2  \\
SwinF. &98.4  &90.8  &72.4  &68.4  &55.1  &77.6  & 74.7  \\ 
\rowcolor{lightblue}EMMA &98.5  &91.1  &74.3  &69.6  &60.1  &79.5  & \underline{76.2}  \\ %\midrule 
 PAIF &98.3  &90.4  &75.0  &66.7  &46.8 & 72.9 &  71.3 \\
MRFS  &98.1  &88.8  &64.4  &65.0  &40.5  &72.4  & 69.1  \\
 SeAF.&98.5  &91.3  &74.7  &69.2  &59.3  &79.9  &  76.1 \\
 SegMiF&98.5  &91.1  &73.5  &68.8  &55.7  &76.5  & 74.7  \\ \midrule 
 \rowcolor{lightred}\Romannum{1}&98.5  &91.2  &74.9  &69.4  &59.7  &80.1  &\textbf{76.2}   \\
\Romannum{2}&98.5  &91.2  &75.5  &69.5  &60.0  &80.6  &76.0   \\
\Romannum{3}&98.5  &91.2  &75.2  &69.6  &59.2  &79.6  &76.2   \\
\Romannum{4}&98.5  &91.0  &74.8  &69.3  &59.2  &80.6  &76.2   \\
\midrule[1.2pt] 
\end{tabular}
}
\caption{Quantitative segmentation (IoU\%↑) on the MSRS dataset. Best and second-best values are \textbf{highlighted} and \underline{underlined}.}
\label{table.MSRS_segmentation}
\end{table}

\begin{figure}[t]
	\centering
	\includegraphics[width=0.45\textwidth]{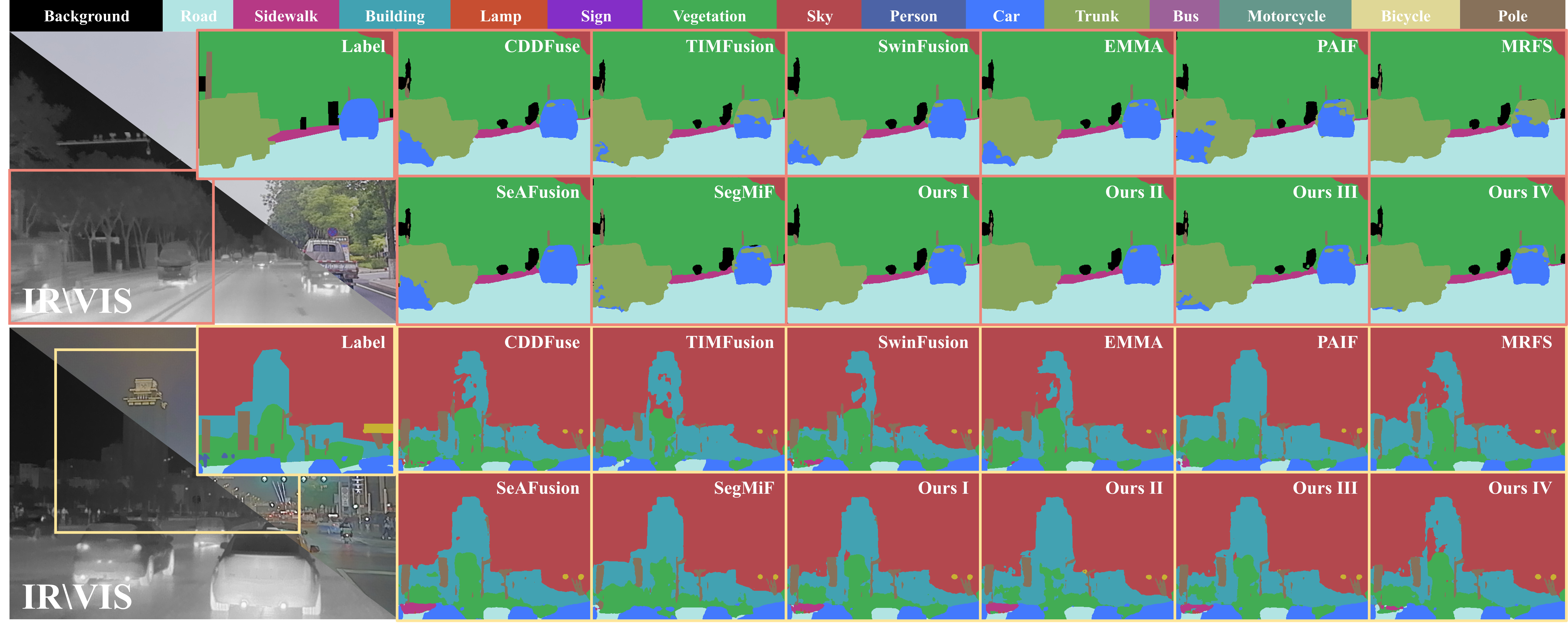}
	\caption{Qualitative segmentation on the FMB dataset.}
	\label{fig:qualitative_seg_fmb}
\end{figure}

\begin{figure}[t]
	\centering
	\includegraphics[width=0.45\textwidth]{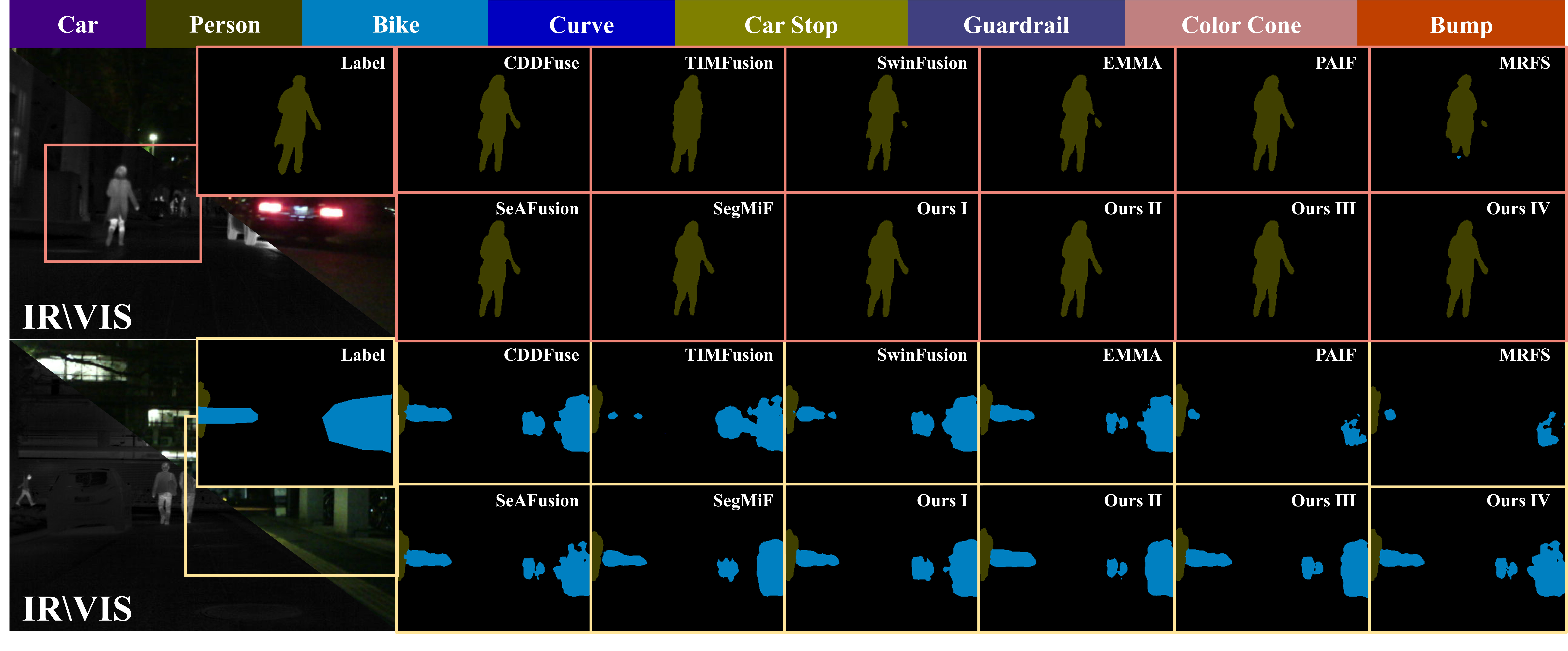}
	\caption{Qualitative segmentation on the MSRS dataset.}
	\label{fig:qualitative_seg_msrs}
\end{figure}

\noindent\textbf{Image fusion metrics.}~To comprehensively evaluate the fusion performance, 7 metrics covering 5 different aspects of fused image quality were selected. These include 2 information theory-based metrics: EN and MI; 1 image feature-based metric: $Q_{abf}$; 2 image structural similarity-based metrics: SSIM and MS-SSIM (MSS); 1 human perception inspired fusion metric: VIF; and 1 color fidelity-based metric: $\Delta E$.~Except for $\Delta E$, higher values for all other metrics indicate superior fusion quality.~The calculation details of these metrics can be found in \cite{ma2019infrared,zhang2020vifb,gaurav2005ciede2000}.

\subsection{Semantic segmentation performance}
\label{sec:segmentation_performance}

\noindent\textbf{Setup.}~To ensure fairness, we use a unified segmentation model to obtain the segmentation results for different fused images.~This unified model is obtained by fine-tuning the segmentation model provided by SegMiF \cite{liu2023segmif} using fused images based on simple average principle. The backbone is Segformer \cite{xie2021segformer}, and this unified model is trained using cross-entropy loss and optimized with AdamW. The fine-tuning process is conducted with a batch size of 8 over 100 epochs, with the first 30 epochs freezing the backbone. The initial learning rate is $5e^{-4}$, which is reduced to $3e^{-5}$ after unfreezing the backbone, following a cosine annealing schedule. An early stopping criterion (patience = 10) is set to prevent over-fitting.~Since the segmentation categories in the test sets of FMB and MSRS differ, we obtained two different unified segmentation models for each dataset through the aforementioned fine-tuning process.~The division of datasets during fine-tuning follow \cite{liu2023segmif, tang2022piafusion}.

\noindent\textbf{Comparison with SOTA methods.}~We present quantitative segmentation results in Tables \ref{table.FMB_segmentation} and \ref{table.MSRS_segmentation}, in which our MultiTaskVIF attains the highest mIoU across both two datasets, demonstrating that the fusion model trained by our MultiTaskVIF framework can learn more semantic information and effectively improve the segmentation performance of fused images.~Furthermore, we provide visualizations in Fig. \ref{fig:qualitative_seg_fmb} and \ref{fig:qualitative_seg_msrs} to further demonstrate our MultiTaskVIF's better segmentation performance in both datasets.

\subsection{Image fusion performance}
\label{sec.fusion_performance}
%\noindent\textbf{Qualitative comparison.}
We present quantitative fusion results on FMB and MSRS datasets in Tables \ref{table.FMB_fusion} and \ref{table.MSRS_fusion}. Our method has excellent performance on almost all metrics, demonstrating its effectiveness in integrating thermal and visible information. Especially, MultiTaskVIF excels in MI and VIF, preserving detailed textures and high-frequency information while maintaining high SSIM, ensuring structural consistency. Additionally, it minimizes color distortion ($\Delta E$) and suppresses fusion artifacts. These results confirm the robustness of MultiTaskVIF in producing clear, informative, and perceptually superior fused images.~Due to space constraints, qualitative results are presented in the supplementary material.

\begin{table}[t]
\centering
\resizebox{0.45\textwidth}{!}{
\begin{tabular}{c|ccccccc}
%\midrule[1.2] 
\hline 
FMB &EN&MI  &VIF &$Q_{abf}$&SSIM&MSS&$\Delta E$\\
\midrule 
CDDF. & \cellcolor{lightred}\textbf{6.78} & 4.15  & \cellcolor{lightblue}\underline{0.87} & 0.67 & \cellcolor{lightblue}\underline{1.00} & 1.06 &6.11 \\  
TIMF. & 6.49 & 3.12  & 0.56 & 0.54 & 0.81 & 0.77 &\cellcolor{lightblue}\underline{4.65} \\
SwinF. & 6.53 & 3.85  & 0.77 & 0.65 & 0.96 & 1.00 &6.22 \\ 
EMMA & 6.77 & 3.95  & 0.83 & 0.64& 0.90 & 1.03  &5.50\\ %\midrule 
 PAIF & 6.58& 3.41 & 0.59 & 0.36 & 0.93 & 1.04  &18.1\\
MRFS  & \cellcolor{lightblue}\underline{6.78} & 3.46  & 0.76 & 0.62 & 0.92 & 0.98 &7.16 \\
 SeAF.& 6.75 & 3.88  & 0.80 & 0.65 & 0.97 & 1.08 &6.17\\
 SegMiF& 6.51 & 3.01  & 0.61 & 0.43 & 0.91 & 1.04 &14.9\\ \midrule 
 \Romannum{1}& 6.63 & \cellcolor{lightred}\textbf{4.40}  & \cellcolor{lightred}\textbf{0.88} & \cellcolor{lightred}\textbf{0.69} & 0.99 & 1.05 &\cellcolor{lightred}\textbf{4.35}\\
\Romannum{2}& 6.63 & 3.93  & 0.77 & 0.65 & 0.95 & \cellcolor{lightblue}\underline{1.07} &5.14\\
\Romannum{3}& 6.61 & 3.99  & 0.76 & 0.66 & 0.95 & 1.05 &4.81\\
\Romannum{4}& 6.65 & \cellcolor{lightblue}\underline{4.15}  & 0.85 & \cellcolor{lightblue}\underline{0.68} & \cellcolor{lightred}\textbf{1.00} & \cellcolor{lightred}\textbf{1.07} &5.58 \\
%\midrule[1.2] 
\hline 
\end{tabular}
}
\caption{Quantitative fusion on the FMB dataset. Best and second-best values are \textbf{highlighted} and \underline{underlined}.}
\label{table.FMB_fusion}
\end{table}

\begin{table}[t]
\centering
\resizebox{0.45\textwidth}{!}{
\begin{tabular}{c|ccccccc}
%\midrule[1.2] 
\hline 
MSRS &EN&MI  &VIF &$Q_{abf}$&SSIM&MSS&$\Delta E$\\
\midrule 
CDDF. & \cellcolor{lightblue}\underline{6.70} & \cellcolor{lightred}\textbf{4.71}  & \cellcolor{lightred}\textbf{1.03} & \cellcolor{lightblue}\underline{0.68} & 0.98 & 1.02 &3.08 \\  
TIMF. & \cellcolor{lightred}\textbf{6.92} & 2.77  & 0.63 & 0.43 & 0.56 & 0.73 &16.3 \\
SwinF. & 6.43 & 3.69  & 0.85 & 0.60 & 0.89 & 0.99 &4.99 \\ 
EMMA & 6.71 & 4.13  & 0.96 & 0.63 &0.94  & 1.03 &3.14 \\ %\midrule 
 PAIF & 5.73 & 2.01  & 0.67 & 0.35 &0.87  & 1.00 &15.8 \\
MRFS  & 6.51 & 2.43   & 0.65 & 0.47 &0.75  & 0.89 &7.30 \\
 SeAF.& 6.65 & 4.03  & 0.97 & 0.67 & 0.99 & 1.05 &\cellcolor{lightred}\textbf{2.48} \\
 SegMiF& 6.08 & 2.22  & 0.62 & 0.37 & 0.80 & 0.97 &10.2 \\ \midrule 
 \Romannum{1}&6.66 &\cellcolor{lightblue}\underline{4.41}    &\cellcolor{lightblue}\underline{1.01}  &\cellcolor{lightred}\textbf{0.68}  &\cellcolor{lightred}\textbf{1.03}  &1.05  &\cellcolor{lightblue}\underline{2.56}  \\
\Romannum{2}&6.65  &3.76    &0.91  & 0.62 &0.99  &\cellcolor{lightblue}\underline{1.05}   &3.41   \\
\Romannum{3}&6.64  &3.75    &0.91  &0.64  &0.99  &\cellcolor{lightred}\textbf{1.05}  &3.33   \\
\Romannum{4}&6.65  &4.13    &0.99  &0.66  &\cellcolor{lightblue}\underline{1.03}  &1.05  &2.96 \\
%\midrule[1.2] 
\hline 
\end{tabular}
}
\caption{Quantitative fusion on the MSRS dataset. Best and second-best values are \textbf{highlighted} and \underline{underlined}.}
\label{table.MSRS_fusion}
\end{table}

\begin{table}[t]
\centering
\resizebox{0.45\textwidth}{!}{
\begin{tabular}{c|cc|cccc|c}
%\midrule[1.2] 
\hline 
\multicolumn{8}{c}{\textbf{Ablation experiment \romannum{1}}}\\ \hline 
Model &Backbone&Decoder&MI &$Q_{abf}$&SSIM& $\Delta E$ &mIoU\%\\
%\midrule 
\hline 
SwinF. &SwinF. & CNN&3.85 & 0.65 & 0.96&6.22 & 64.6 \\  
\Romannum{1}&SwinF. & MTH & \cellcolor{lightred}\textbf{4.40} & \cellcolor{lightred}\textbf{0.69} & \cellcolor{lightred}\textbf{0.99}&\cellcolor{lightred}\textbf{4.35} & \cellcolor{lightred}\textbf{65.7}   \\ \hline %\midrule 
EMMA&Ufuser & CNN&\cellcolor{lightred}\textbf{3.95} & 0.64 & 0.90&5.50 & 64.7  \\  
\Romannum{2}&Ufuser & MTH & 3.93 & \cellcolor{lightred}\textbf{0.65} & \cellcolor{lightred}\textbf{0.95}&\cellcolor{lightred}\textbf{5.14} & \cellcolor{lightred}\textbf{65.1}   \\ \hline % \midrule 
SeAF.n&SeAF. & CNN&3.88 & 0.65 & \cellcolor{lightred}\textbf{0.97}&6.17 & 65.2  \\  
\Romannum{3}&SeAF. & MTH & \cellcolor{lightred}\textbf{3.99} & \cellcolor{lightred}\textbf{0.66} & 0.95 &\cellcolor{lightred}\textbf{4.81}& \cellcolor{lightred}\textbf{65.6}   \\ \hline %\midrule 
SegMiF &SegMiF& CNN&3.01 & 0.43 & 0.91&14.9 & 65.4  \\  
\Romannum{4}&SegMiF & MTH & \cellcolor{lightred}\textbf{4.15} &  \cellcolor{lightred}\textbf{0.68} &  \cellcolor{lightred}\textbf{1.00} & \cellcolor{lightred}\textbf{5.58}&  \cellcolor{lightred}\textbf{65.8}  \\
%\midrule[1.2]
\hline 
\multicolumn{8}{c}{\textbf{Ablation experiment \romannum{2}~to \romannum{5}~(based on model \Romannum{4})}}\\ \hline 
Exp. &\multicolumn{2}{c|}{Configurations}&MI &$Q_{abf}$&SSIM& $\Delta E$ &mIoU\%\\ \midrule
\romannum{2}&\multicolumn{2}{c|}{w/o HIA-F}& 3.98 & 0.66 & 0.98&4.50 & 65.1  \\
\romannum{3}&\multicolumn{2}{c|}{w/o $\mathcal{L}_{seg}$}& 3.98 & 0.67 & 0.99 &5.27 &  65.2 \\
\romannum{4}&\multicolumn{2}{c|}{w/o $\mathcal{L}_{color}$}& 4.18 & 0.68 &  0.99&6.08 &  65.3 \\
\romannum{5}&\multicolumn{2}{c|}{Channel size: 3→1}& \cellcolor{lightred}\textbf{4.35} & 0.67 & 0.95 &\cellcolor{lightred}\textbf{4.24} & 64.6  \\ \hline % \midrule
Ours&\multicolumn{2}{c|}{w/ all}& 
4.15 & \cellcolor{lightred}\textbf{0.68} & \cellcolor{lightred}\textbf{1.00} &\textbf{5.58}& \cellcolor{lightred}\textbf{65.8}  \\
%\midrule[1.2]
\hline 
\end{tabular}
}
\caption{Ablation experiment results on the testset of FMB.}% Best values are \textbf{highlighted}.}
\label{table.Aba}
\end{table}

\subsection{Ablation studies}

To evaluate the rationality of MultiTaskVIF, we conduct ablation studies on the FMB testset. MI, $Q_{abf}$, SSIM, $\Delta E$ and mIoU are used to quantitatively validate the fusion and segmentation effectiveness. The results of experimental groups are shown in Table \ref{table.Aba}.

\noindent\textbf{Multi-task Head.}~In Exp.\unskip~\romannum{1}, to verify the effectiveness of MTH, we replaced the model's decoder with a CNN. However, due to the inability to maintain consistent loss functions and training settings across different network architectures, we directly compared our models with existing VIF models that share the same backbone as ours in the actual experiments. The results demonstrate that our designed MTH can effectively enhance fused image reconstruction, enabling the fused images to preserve more texture details and color information while maintaining high structural consistency. Moreover, MTH effectively incorporates the semantic information of fused features during fused image reconstruction, leading to superior segmentation performance of fused images. Consistent good results with different backbones further demonstrate the flexibility of MTH and the generalizability of MultiTaskVIF.

\noindent\textbf{Improved HIA-F.}~In Exp.\unskip~\romannum{2}, we eliminated HIA-F, as introduced in Section \ref{Sec.MTH}, and observed that without HIA-F, the multi-task learning network structure alone struggled to effectively integrate semantic information into the fusion branch, leading to suboptimal segmentation performance of the fused images. Moreover, injecting more semantic information during the fusion reconstruction process further contributes to enhancing the texture details and structural consistency of the fused images.

\noindent\textbf{Terms in loss function.}~Then, in Exp.\unskip~\romannum{3}, we eliminate the segmentation loss $\mathcal{L}_{seg}$ in Eq. \ref{eq.total}. The results of Exp. \romannum{3}~demonstrate that $\mathcal{L}_{seg}$ can effectively guide the integration of more semantic information into the model training, thereby enhancing the segmentation performance of the fused images. In Exp.\unskip~\romannum{4}, we eliminate the last term in Eq. \ref{eq.fusion}, which is the color-preserving term $\mathcal{L}_{color}$. The Exp. \romannum{4}'s results indicate that $\mathcal{L}_{color}$ is necessary for minimizing color distortion, and thus improving the segmentation performance of fused images.

\noindent\textbf{Three channel input.}~Finally, in Exp.\unskip~\romannum{5}, we change the model's input and output data from three channels to a single channel.~The three-channel method fuses all RGB channels to directly generate the fused RGB image. In contrast, the single-channel method fuses only the grayscale (Y) channel, then combines it with the Cb and Cr channels of visible image.~Although single-channel fusion is a common approach in VIF, we argue that color information also contributes to enhancing image fusion, particularly in segmentation-oriented fusion tasks, where color information theoretically facilitates the extraction of segmentation features \cite{rother2004grabcut}. The results of Exp.\unskip~\romannum{5}~further validate our hypothesis. While single-channel fusion preserves more original image information, three-channel fusion can more effectively utilize these information, significantly improving the segmentation performance of the fused images and achieving superior visual similarity and image quality. Thus, we conclude that three-channel fusion is more suitable for segmentation-oriented VIF tasks.

In summary, ablation results in Table \ref{table.Aba} demonstrate the effectiveness and rationality of our proposed methods.

\subsection{Computational efficiency}

\begin{table}[ht]
\centering
\renewcommand\arraystretch{1}
\resizebox{0.45\textwidth}{!}{
\begin{tabular}{c|c|ccc}
%\midrule [1.2]
\hline 
Model &Input size &Params(M)&FLOPs(G)&Mem usage(MB)\\
\midrule 
SwinF. &(256,256,1) & 1.71&124.03 & 686.22 \\  
\Romannum{1}&(256,256,3) & 2.47&125.26 & 689.65\\ \midrule 
EMMA&(256,256,1) & 80.31&118.08& 353.37\\  
\Romannum{2}&(256,256,3) & 2.13&49.08 & 167.00 \\ \midrule 
SeAF.&(256,256,1) & 13.06&13.98 &4063.57  \\  
\Romannum{3}&(256,256,3) &0.93&60.84&245.68  \\ \midrule 
SegMiF&(256,256,1) & 45.25&51.20& 430.90  \\  
\Romannum{4}&(256,256,3) & 1.28&83.83&214.52  \\
%\midrule [1.2]
\hline 
\end{tabular}
}
\caption{Computational efficiency during training.}
\label{table.Compu}
\end{table}

Table \ref{table.Varients} has shown that MultiTaskVIF only requires one model and one stage for training. In this section, we further compare the computational efficiency during training of MultiTaskVIF with other methods. As detailed in Table \ref{table.Compu}, we quantify the number of parameters, the FLOPs, and the memory usage (tested on images with the size of $256\times256$) for MultiTaskVIF \Romannum{1} to \Romannum{4} and other VIF methods with same backbone as us. Obviously, compared to methods that require multiple models for training (EMMA, SeAFusion, and SegMiF), our method has lower parameter and memory usage. This means that under limited memory resources, our method can support a larger batch size and crop size. Compared to methods that involve only a single model for training (such as SwinFusion), our method (using 3-channel image as input) achieves superior fusion and segmentation performance with only a minimal increase in computational cost. Additionally, the single-stage training process simplifies the overall training procedure, reduces the number of hyperparameters that need to be tuned, and avoids the potential cumulative errors introduced by multi-stage training.

\subsection{Gradient Analysis}

\begin{figure}[ht]
	\centering
	\includegraphics[width=0.45\textwidth]{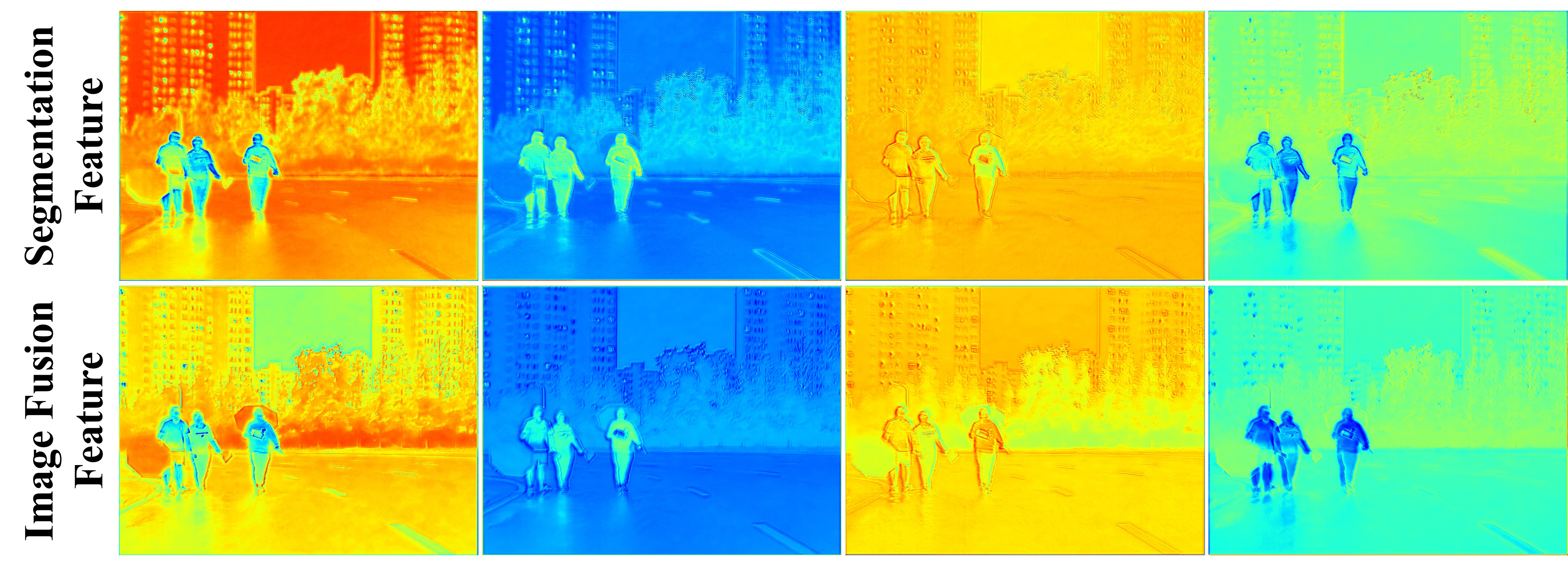}
	\caption{Feature maps of the two branches within MTH.}
	\label{fig:feature_map}
\end{figure}

Multi-task learning relies on task correlation for mutual enhancement. To verify the correlation between VIF and segmentation tasks, we applied the gradient projection method to assess their influence on each other. During MultiTaskVIF \Romannum{4}'s training, we recorded the gradients of the fusion and segmentation losses with respect to their shared parameters, along with their gradient projections. After 100 epochs, the average projection of the fusion task onto the segmentation task was $1.9e^{-1}$, and the projection of the segmentation task onto the fusion task was $2.7e^{-3}$. Both positive values confirm that the segmentation branch positively contributes to fusion model training. Fig. \ref{fig:feature_map} shows the intermediate feature maps from MTH, further illustrating the correlation between fusion and segmentation features. This explains why MTL effectively enhances both fusion and segmentation performance in VIF models.

\section{Conclusions}
In this paper, we propose MultiTaskVIF, a general training framework for segmentation-oriented visible-infrared image fusion. Inspired by multi-task learning, MultiTaskVIF adopts a concise parallel training framework to enhance both fusion quality and downstream application performance. To achieve this, we design a dual-branch multi-task decoder (MTH) for simultaneous fusion reconstruction and semantic segmentation. Extensive experiments validate the effectiveness of our MultiTaskVIF training framework and the rationality of each module design, providing a solid foundation for future application-oriented VIF research.

{\small
\bibliographystyle{ieeenat_fullname}
\bibliography{main}
}

\end{document}